\documentclass{article}

\usepackage[preprint, nonatbib]{neurips_2020}

\usepackage[utf8]{inputenc} 
\usepackage[T1]{fontenc}    
\usepackage{hyperref}       
\usepackage{url}            
\usepackage{booktabs}       
\usepackage{amsfonts}       
\usepackage{nicefrac}       
\usepackage{microtype}      
\usepackage{graphicx}       
\usepackage{subcaption}

\title{CancerNet-SCa: Tailored Deep Neural Network Designs for Detection of Skin Cancer from Dermoscopy Images}

\author{
  James Ren Hou Lee$^{1*}$, Maya Pavlova$^{1,3*}$, Mahmoud Famouri$^{3}$, and Alexander Wong$^{1,2,3}$\\
  $^{1}$ Vision and Image Processing Research Group, University of Waterloo, Waterloo, ON, Canada\\
  $^{2}$ Waterloo Artificial Intelligence Institute, University of Waterloo, Waterloo, ON, Canada\\
  $^{3}$ DarwinAI Corp., Waterloo, ON, Canada \\
  \texttt{$^{*}$ equal authors} \\
}

\begin{document}

\maketitle

\begin{abstract}

Skin cancer continues to be the most frequently diagnosed form of cancer in the U.S., with not only significant effects on health and well-being but also significant economic costs associated with treatment.  A crucial step to the treatment and management of skin cancer is effective skin cancer detection due to strong prognosis when treated at an early stage, with one of the key screening approaches being dermoscopy examination. Motivated by the advances of deep learning and inspired by the open source initiatives in the research community, in this study we introduce CancerNet-SCa, a suite of deep neural network designs tailored
for the detection of skin cancer from dermoscopy images that is open source and available to the general public as part of the Cancer-Net initiative. To the best of the authors' knowledge, CancerNet-SCa comprises of the first machine-designed deep neural network architecture designs tailored specifically for skin cancer detection, one of which possessing a self-attention architecture design with attention condensers.  Furthermore, we investigate and audit the behaviour of CancerNet-SCa in a responsible and transparent manner via explainability-driven model auditing. While CancerNet-SCa is not a production-ready screening solution, the hope is that the release of CancerNet-SCa in open source, open access form will encourage researchers, clinicians, and citizen data scientists alike to leverage and build upon them.

\end{abstract}

\section{Introduction}
Skin cancer is the most frequently occurring form of cancer in the U.S., with over 5 million new cases diagnosed each year~\cite{cancerstats}, and continues to rise with each year that passes.  Furthermore, the annual cost of treating skin cancer in the U.S. alone is estimated to be over \$8 billion~\cite{cost}.  Fortunately, prognosis is good for many forms of skin cancer when detected early~\cite{cancerstats2}, and as such early skin cancer detection becomes very important for patient recovery.  This is particularly critical for melanoma, the most deadly form of skin cancer that, if left undiagnosed and untreated at an early stage,  can spread beyond its original location to nearby skin and organs until surgery is no longer sufficient and treatment methods such as radiation are required \cite{matthews2017epidemiology}. As such, early diagnosis and preventative measures are especially critical for melanoma as the death rate and cost of treatment both increase drastically as the cancer progresses from Stage I to Stage IV \cite{glaister2013automatic}. However, if melanoma is diagnosed early on, a simple surgery to remove the lesion can increase survival rates by stopping the cancer from spreading beyond its origin \cite{celebi2007methodological, celebi2015state}.

Currently, the most popular method of skin lesion diagnosis is the dermoscope assisted method \cite{braun2004dermoscopy}, which is able to achieve a diagnostic accuracy of roughly 75\% to 97\% \cite{celebi2005unsupervised}. However, it was also found that in the hands of a dermatologist that has limited experience with the instrument, the use of a dermoscope may reduce the diagnostic accuracy rather than augmenting it. In addition, the ``lack of reproducibility and subjectivity of human interpretation'' \cite{celebi2015state} associated with human based diagnosis is one of the main reasons why there has been a significant increase in interest for computer assisted diagnosis of skin cancer. The use of computer aids for the diagnosis of pigmented skin lesions has been shown to be accurate and practical, and can improve biopsy decision making \cite{hoffmann2003diagnostic}, as well as act as a pre-screening tool to reduce the amount of a time a professional spends on each case.

Motivated by the tremendous advances in the field of deep learning~\cite{lecun2015deep} and the great potential it has shown in a wide range of clinical decision support applications~\cite{Arcadu,wang2020covidnet,gunraj2020covidnetct,wong2020covidnets}, a number of recent studies have explored the efficacy of deep neural networks for the purpose of skin cancer detection~\cite{9034624,Esteva}.  For example, in the recent study by Budhiman et. al.~\cite{9034624}, a comprehensive exploration of different residual network architectures was conducted for the purpose of melanoma detection, with the best quantitative results found when leveraging a ResNet-50 \cite{he2016deep} architecture.

Motivated by the challenge of skin cancer detection, and inspired by the open source and open access efforts of the research community, in this study we introduce \textbf{CancerNet-SCa}, a suite of deep neural network designs tailored for the detection of skin cancer from dermoscopy images, one of which possessing a self-attention architecture design with attention condensers~\cite{wong2020tinyspeech,wong2020attendnets}.  To construct CancerNet-SCa, we leveraged a machine-driven design strategy that leverages human experience and ingenuity with the meticulousness and raw speed of machines. We further leverage the International Skin Imaging Collaboration (ISIC) dataset~\cite{isic2020dataset} for this study, and illustrate the efficacy of CancerNet-SCa when compared to previously proposed deep neural network architectures~\cite{he2016identity}.  To the best of the authors' knowledge, CancerNet-SCa comprises of the first machine-designed deep neural network architecture designs tailored specifically for skin cancer detection.  CancerNet-SCa is available to the general public in an open-source and open access manner as part of the Cancer-Net initiative\footnote{https://github.com/jamesrenhoulee/CancerNet-SCa}, an open initiative launched as a complement to the COVID-Net initiative~\cite{wang2020covidnet}.  While CancerNet-SCa is not a production-ready screening solution, the hope is that the release of CancerNet-SCa will encourage researchers, clinicians, and citizen data scientists alike to leverage and build upon them.

The paper is organized as follows.  Section~\ref{methods} describes the methodology leveraged to build the proposed CancerNet-SCa, the overall network architecture designs of CancerNet-SCa, as well as the explainability-driven performance validation strategy leveraged to audit CancerNet-SCa.  Section~\ref{results} presents the quantitative results for evaluating the efficacy of CancerNet-SCa, qualitative results to gain insights into the decision-making behaviour of CancerNet-SCa, and a discussion on the broader impact of methods such as CancerNet-SCa for aiding the clinical decision support process. Finally, conclusions are drawn and future work discussed in Section~\ref{conclusion}.

\begin{figure}
  \centering
  \includegraphics[width = 0.8\linewidth]{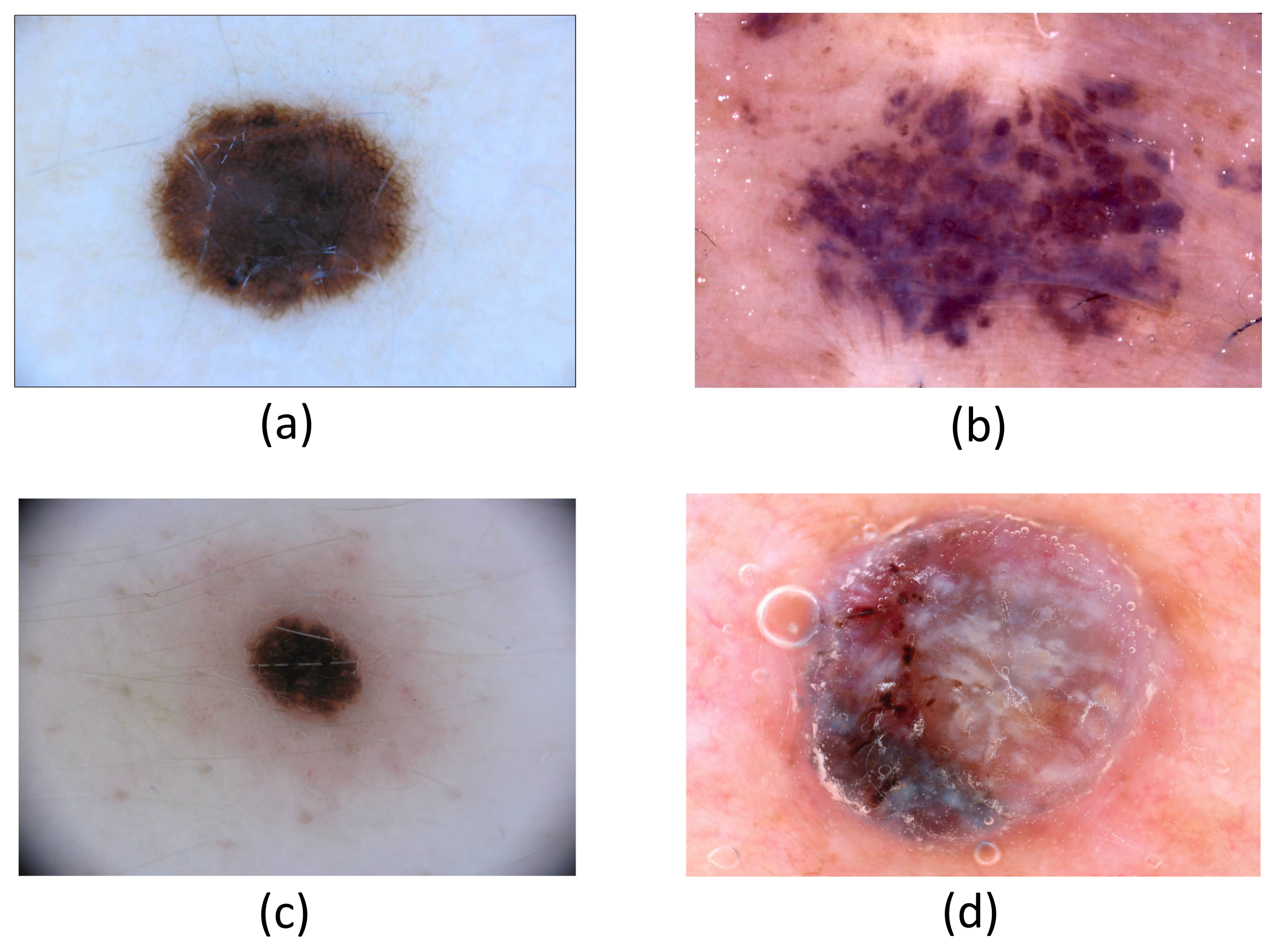}
  \caption{Sample images from the ISIC Dataset leveraged to build CancerNet-SCa. Dermoscopy images (a) and (b) are both benign, while images (c) and (d) are both malignant. Image (b) can easily be mistaken for a malignant lesion, while image (c) can easily be misclassified as benign to the untrained eye.}
  \label{fig:isic_examples}
\end{figure}

\begin{figure*}[t]
  \centering
  \includegraphics[width= \linewidth]{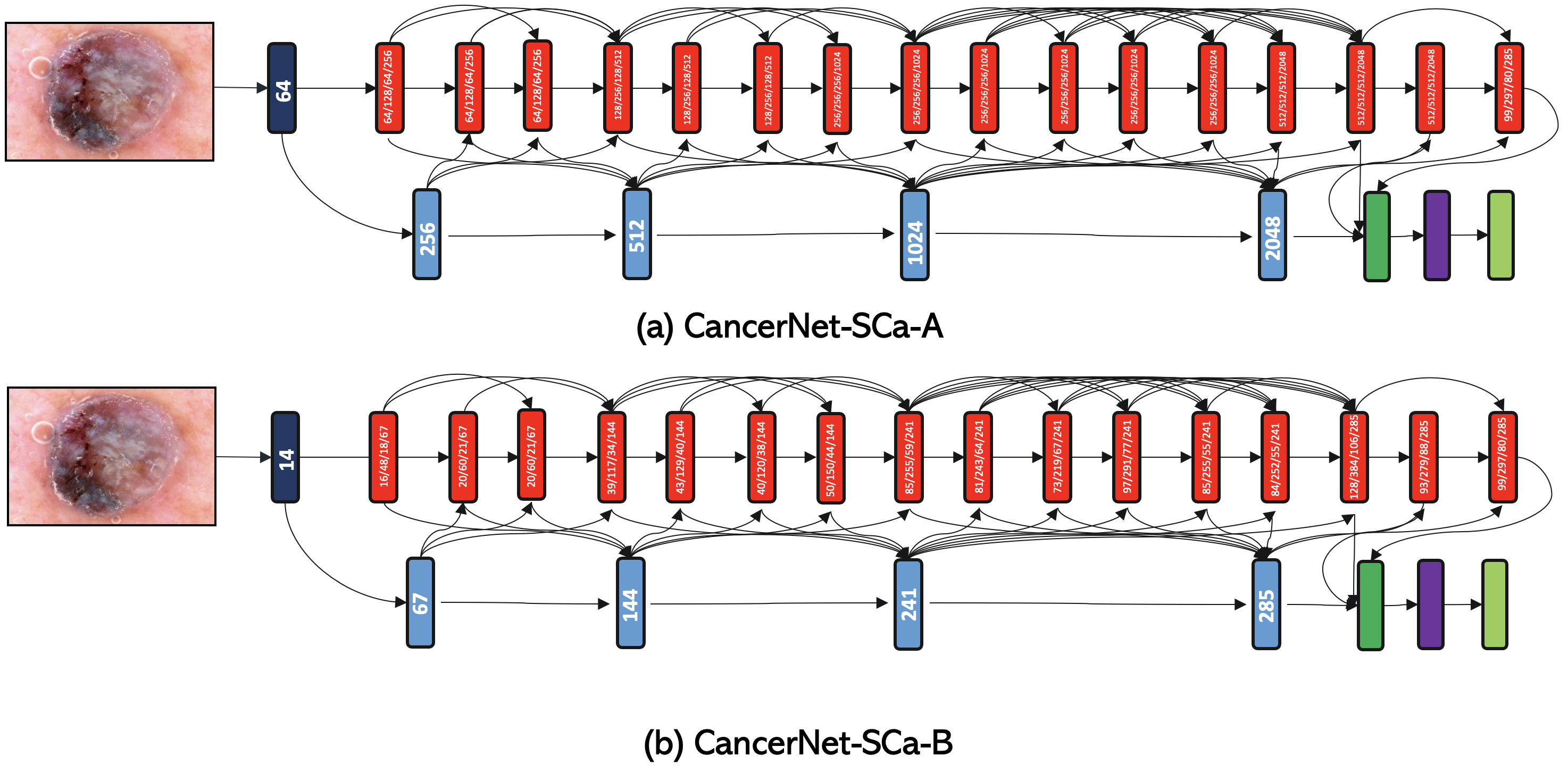}
  \includegraphics[width= \linewidth]{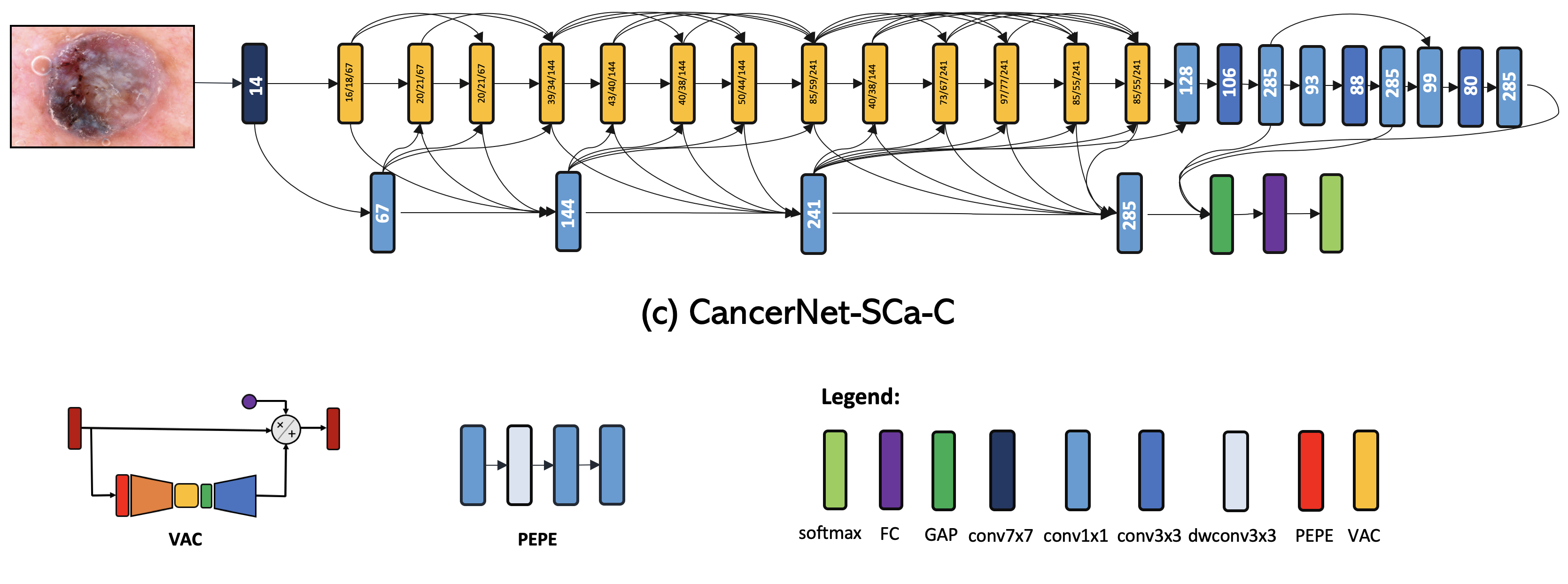}
  \caption{The proposed CancerNet-SCa network architectures.  The number in each convolution module represents the number of channels. The numbers in each visual attention condenser represents the number of channels for the down-mixing layer, the embedding structure, and the up-mixing layer, respectively (details can be found in~\cite{wong2020attendnets}).  It can be observed that all CancerNet-SCa architectures exhibit both great macroarchitecture and microarchitecture design diversity as well as lightweight macroarchitecture design characteristics such as attention condenser design patterns and projection-expansion-projection-expansion (PEPE) design patterns comprised of channel dimensionality reduction, depthwise convolutions, and pointwise convolutions.}
  \label{fig:architecture}
\end{figure*}

\section{Methods}
\label{methods}
\subsection{Data preparation}
\label{dataprep}
In this study, we leverage the International Skin Imaging Collaboration (ISIC) dataset, which is an open source public access archive of skin images. The dataset consists of 23,906 dermoscopy images at the time of the study, comprising a variety of skin cancer types such as Squamous Cell Carcinoma, Basal Cell Carcinoma, and Melanoma. In the dataset, a total of 21,659 dermoscopy images were labelled as either benign or malignant, and we leveraged this set to build CancerNet-SCa. Random partitioning was conducted to form the training, validation, and test sets, with the test set consisting of a balanced split of 221 benign samples and 221 malignant samples out of the total set of dermoscopy images.  The balanced split enables for a better assessment of performance given the imbalances in the ISIC dataset.  During training, data augmentation was applied which included rotations (up to 30 degrees), shifts (up to 10 percent), and vertical and horizontal flipping. Each image was resized to a size of 224x224 pixels, with the 3-channel color information retained. Apart from these standard transformations, no additional pre-processing was performed for training.

\subsection{Machine-driven design exploration}

To build the proposed CancerNet-SCa deep neural network designs to be as tailored as possible around the task of skin cancer detection, we leveraged a machine-driven design exploration strategy to automatically design a highly customized deep neural network architecture based on characteristics of the ISIC dataset.  More specifically, the machine-driven design exploration strategy we leveraged was generative synthesis~\cite{wong2018ferminets}, where  the problem of identifying a specific deep neural network architecture tailored for a specific task is formulated as a constrained optimization problem, using a universal performance function~\cite{wong2018netscore} and a set of operational constraints related to the task and data at hand.  This constrained optimization problem is solved through an iterative process, with an initial design prototype to initialize the optimization process. In this study, the operational constraint leveraged was that the validation accuracy of the designed deep neural network exceeded that of the ResNet-50 architecture leveraged in~\cite{9034624}, which was found by the authors of that study to provide the best quantitative results amongst different residual network architectures.  For the initial design prototype, residual architecture design principles \cite{he2016identity} were leveraged in this study. The use of residual connections have been shown to alleviate vanishing gradient and dimensionality problems, allowing networks to learn faster and easier with minor additional cost to architectural or computational complexity.  Furthermore, given the iterative nature of the machine-driven design exploration strategy, we selected three of the generated deep neural network architecture designs to construct CancerNet-SCa (i.e., CancerNet-SCa-A, CancerNet-SCa-B, and CancerNet-SCa-C).

\subsection{Network architecture}

The proposed CancerNet-SCa architectures are shown in Figure \ref{fig:architecture}.  A number of interesting observations can be made about the CancerNet-SCa architectures.  First of all, it can be observed that the macroarchitecture designs exhibited in all CancerNet-SCa network architectures are highly diverse and heterogeneous, with a mix of spatial convolutions, pointwise convolutions, and depthwise convolutions, all with different microarchitecture designs.  This reflects the fact that a machine-driven design exploration strategy was leveraged and allows for very fine-grained design decisions to be made to best tailor for the task of skin cancer detection.  Second, it can be observed that very lightweight design patterns are exhibited in the proposed CancerNet-SCa architectures.  For example, CancerNet-SCa-A and CancerNet-SCa-B leverage projection-expansion-projection-expansion (PEPE) design patterns comprised of channel dimensionality reduction, depthwise convolutions, and pointwise convolutions.  In another example, CancerNet-SCa-C exhibits an efficient self-attention architecture design that leverages attention condensers~\cite{wong2020attendnets,wong2020tinyspeech}, efficient self-attention mechanisms that produce condensed embedding characterizing joint local and cross-channel activation relationships and performs selective attention accordingly to improve representational capability.  This is important as it better facilitates for real-time diagnosis, potentially on digital dermoscopy scanners.  The unique designs of CancerNet-SCa thus illustrate the benefits of leveraging machine-driven design exploration to create deep neural network architectures tailored to skin cancer detection.

\subsection{Explainability-driven performance validation}

To audit CancerNet-SCa in a responsible and transparent manner, we take inspiration from~\cite{wang2020covidnet} and conducted an explainability-driven performance validation by leveraging GSInquire~\cite{lin2019explanations}, an explainability method that has been shown to provide state-of-the-art quantitative interpretability performance in a way that reflects the decision-making process of the underlying deep neural network.  This allows us to not only ensure that decisions made by CancerNet-SCa are not based on erroneous visual cues, but also gain better insight into its decision-making behaviour to better understand what visual cues may be important for detecting skin cancer.

    \section{Results and Discussion}
    \label{results}
\subsection{Quantitative results}

We evaluate the performance of the proposed CancerNet-SCa deep neural network designs using the test set of dermoscopy images described in Section~\ref{dataprep}.  Furthermore, for evaluation purposes, we compare it with the 50-layer residual deep convolutional neural network architecture~\cite{he2016identity}, which was leveraged by Budhiman et al.~\cite{9034624} to achieve the best quantitative results in their experiments.  The performance metrics leveraged in this study are accuracy, sensitivity, and positive predictive value (PPV).  Construction and evaluation are conducted using TensorFlow, with tested architectures trained using Adam optimizer, with LR=0.0001, epochs=80, momentum=0.9, and batch rebalancing.

The results are shown in Table \ref{tab:quant_results} and Table~\ref{tab:quant_results2}. A number of observations can be made. First, it can be observed that all proposed CancerNet-SCa designs achieved improved accuracy when compared to the ResNet-50 network architecture while achieving significantly reduced architectural and computational complexity.  For example, CancerNet-SCa-B achieved 6.1\% higher accuracy when compared to the ResNet-50 network architecture while possessing 29.5$\times$ fewer parameters and requiring $\sim$18$\times$ fewer FLOPs.  This illustrates the benefits of leveraging a machine-driven design exploration strategy for designing a deep neural network architecture that finds a strong balance between efficiency and accuracy.  Second, it can be observed that all proposed CancerNet-SCa designs achieved higher sensitivity and positive predictive value (PPV) than that achieved with the ResNet-50 network architecture.  For example, CancerNet-SCa-A achieved 14.1\% higher sensitivity and 0.5\% higher PPV when compared to the ResNet-50 network architecture.  This further illustrates the benefits of leveraging a machine-driven design exploration strategy for designing a highly customized deep neural network architecture tailored specifically for skin cancer detection.  Finally, we can observe that the CancerNet-SCa designs have different performance-efficiency tradeoffs, with CancerNet-SCa-A providing the highest sensitivity, CancerNet-SCa-B having the lowest architectural complexity, highest accuracy, and highest PPV, and CancerNet-SCa-C having the lowest computational complexity and highest PPV. This illustrates how a machine-driven design exploration strategy can allow for greater flexibility to meet the requirements of the use case (e.g., on-device examination vs. cloud-driven examination).

\begin{table}[]
\caption{Comparison of parameters, FLOPs, and accuracy for tested network architectures on the ISIC dataset. Best results highlighted in \textbf{bold}.}
\begin{center}
\begin{tabular}{|c|c|c|c|}
\hline\\[-0.85em]
Architecture & Params (M) & FLOPs (G) & Accuracy (\%)\\[3pt]
\hline
ResNet-50 \cite{9034624} & 23.52 & 7.72 & 78.3\\
\hline
CancerNet-SCa-A & 13.65 & 4.66 & 83.7\\
\hline
CancerNet-SCa-B & \textbf{0.80} & 0.43 & \textbf{84.4}\\
\hline
CancerNet-SCa-C & 1.19 & \textbf{0.40} & 83.9\\
\hline
\end{tabular}\par
\bigskip
\label{tab:quant_results}
\end{center}
\vspace{-0.2in}
\end{table}

\begin{table}[]
\caption{Sensitivity and positive predictive value (PPV) comparison on the ISIC dataset. Best results highlighted in \textbf{bold}.}
\begin{center}
\begin{tabular}{|c|c|c|}
\hline\\[-0.85em]
Architecture & \textbf{Sensitivity ($\%$)} & \textbf{PPV ($\%$)}\\[3pt]
\hline
ResNet-50 \cite{9034624} & 78.7 & 78.0\\
\hline
CancerNet-SCa-A & \textbf{92.8} & 78.5\\
\hline
CancerNet-SCa-B & 91.4 & \textbf{80.2}\\
\hline
CancerNet-SCa-C &  90.0 & \textbf{80.2} \\
\hline
\end{tabular}\par
\bigskip
\label{tab:quant_results2}
\end{center}
\vspace{-25pt}
\end{table}

\subsection{Qualitative results}

\begin{figure}
  \centering
  \includegraphics[width= \linewidth]{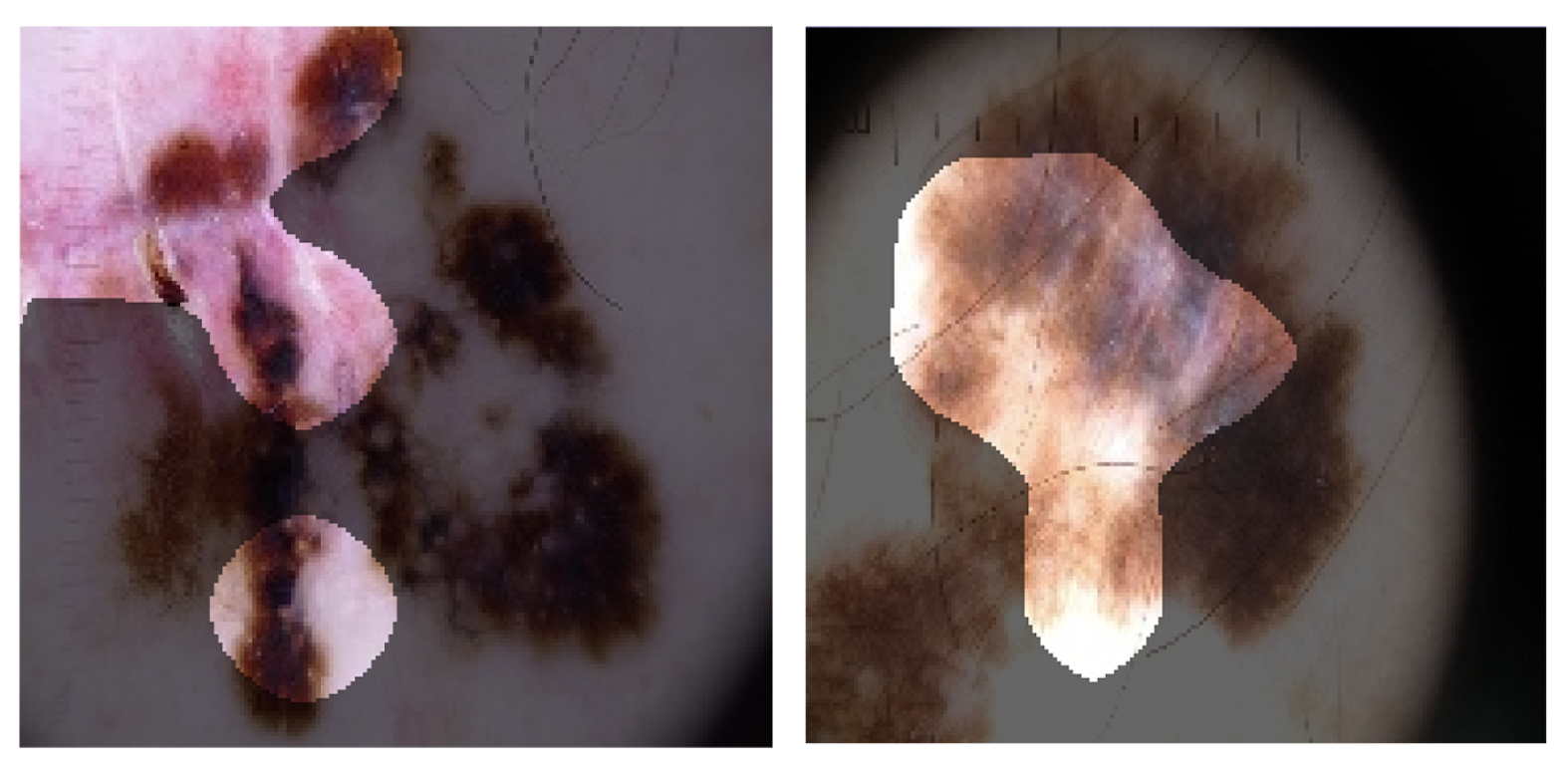}
  \caption{Example skin images from the ISIC dataset and their associated diagnostically relevant imaging features as identified by GSInquire \cite{lin2019explanations}, using CancerNet-SCa-A. The bright regions indicate the imaging features identified to be relevant.}
  \label{fig:gsinquire}
\end{figure}

In order to better understand how CancerNet-SCa makes detection decisions based on dermoscopy images, we leveraged GSInquire \cite{lin2019explanations} for explainability-driven performance validation and insight discovery to audit the decision-making process. Examples of this are shown in Figure \ref{fig:gsinquire}, where it can be seen that CancerNet-SCa-A leverages the color and textural heterogeneities as well as morphological irregularities found within the skin lesions as exhibited in the dermoscopy images to differentiate between benign and malignant skin lesions. As such, such an explainability-driven performance validation helps screen for erroneous decision-making behaviour that rely on irrelevant visual indicators and imaging artifacts, and emphasizes the importance of auditing deep neural networks when designed for clinical applications, as it can increase the trust that practitioners have towards deep learning.

\subsection{Discussion and broader impact}

Skin cancer continues to be a prominent problem for the health and well-being of society, with millions of new cases annually and causing thousands of deaths annually while costing billions of dollars in the United States alone.  Not only are most biopsies unnecessary (only 1 in 36 biopsies yield a case of melanoma \cite{bhattacharya2017precision}), the cost of misdiagnoses and unnecessary biopsies can quickly accumulate. This adds an expensive and needless burden to both the patient and the system, while taking up precious time which could be allocated to treating additional patients.

The benefits of computer-aided strategies such as CancerNet-SCa are two-fold. Not only do they provide dermatologists with valuable second opinions during diagnosis, they also save time by acting as pre-screening tools in the diagnostic process. The goal of CancerNet-SCa is not to replace dermatologists, but instead to aid professionals in their decision-making processes as well as act as a basis for others to improve upon and accelerate advances in this area. When correctly leveraged with professional knowledge, CancerNet-SCa will hopefully impact the field of dermatology in a positive manner. The fact that CancerNet-SCa underwent explainability-driven auditing will hopefully allow for greater trust as well as better understanding of its decision-making behaviour.  As one of the major deterrents of deep learning in the medical field is the ``black-box'' nature of these systems, granting additional insight on how network decisions are reached can result in more trust towards the systems - a crucial first step towards the widespread adoption of artificial intelligence for health and safety.

\section{Conclusion}
\label{conclusion}
\vspace{-0.1in}
In this study, we introduced CancerNet-SCa, a suite of deep neural network designs tailored for the detection of skin cancer from dermoscopy images, each with a different balance in performance and efficiency. Designed via a machine-driven design exploration strategy, CancerNet-SCa is available open source and available to the general public as part of the Cancer-Net initiative. Experimental results using the ISIC dataset show that the proposed CancerNet-SCa designs can achieve strong skin cancer detection performance while providing a strong balance between computational and architectural efficiency and accuracy. An explainability-driven audit of CancerNet-SCa is also conducted, showing that prediction is performed by leveraging relevant abnormalities found within skin lesion images, rather than random visual indicators and imaging artifacts. Given the promise of leveraging a machine-driven design exploration strategy for creating a highly customized deep neural network architecture for cancer detection, we aim to explore this strategy for creating deep neural networks for other forms of cancer such as lung cancer, breast cancer, and prostate cancer and making them publicly available for researchers, clinicians, and citizen data scientists alike to leverage and build upon them.

\section*{Acknowledgments}
\vspace{-0.1in}
We thank the Devon Stoneman Memorial Scholarship, and thank NSERC and the Canada Research Chairs program.

\medskip

\small

\bibliographystyle{IEEEtran}
\bibliography{references}

\end{document}